\documentclass{article}

\usepackage[preprint, nonatbib]{neurips_2019}

\usepackage[utf8]{inputenc} %
\usepackage[T1]{fontenc}    %
\usepackage{hyperref}       %
\usepackage{url}            %
\usepackage{booktabs}       %
\usepackage{amsfonts}       %
\usepackage{nicefrac}       %
\usepackage{microtype}      %

\usepackage{tikz}
\usetikzlibrary{shapes}
\usetikzlibrary{automata,topaths}
\usetikzlibrary{calc}

\usepackage{wrapfig}

\usepackage{caption}
\usepackage{subcaption}

\usepackage{floatrow}

\usepackage{ragged2e}

\usepackage{afterpage}

\usepackage{rotating}
\usepackage{makecell}
\usepackage{enumitem}

\usepackage{mathtools}

\usepackage{snapshot}

\usepackage[toc,page]{appendix}

\usepackage{url}

\usepackage[
backend=biber,
style=ieee,
citestyle=ieee,
]{biblatex}
\AtBeginBibliography{\small}
\let\URL\url
\makeatletter
\def\url{\begingroup \catcode`\%=12\catcode`\#=12\relax\printurl}
\def\printurl#1{\@URL#1;\@nil\endgroup}
\def\@URL#1;#2\@nil{\URL{#1}\ifx\relax#2\relax \else;\ \  \url{#2\relax}\fi}
\makeatother

\newcommand{\imageone}[1]{}
\newcommand{\imagetwo}[1]{}



\usepackage{microtype}

\RequirePackage[none]{hyphenat}
\def\theLetterSpace{-0.5pt}
\def\extraWordSpace{-0.5pt}
\newcommand\spaceout[2][\theLetterSpace]{%
  \def\LocalLetterSpace{#1}\expandafter\spaceouthelpA#2 \relax\relax}
\def\spaceouthelpA#1 #2\relax{%
  \spaceouthelpB#1\relax\relax%
  \ifx\relax#2\else\kern\extraWordSpace\ \kern\LocalLetterSpace\spaceouthelpA#2\relax\fi
}
\def\spaceouthelpB#1#2\relax{%
  #1%
  \ifx\relax#2\else
    \kern\LocalLetterSpace\spaceouthelpB#2\relax%
  \fi
}

\newif\ifdraft
\drafttrue

\ifdraft
\newcommand{\fpc}[1]{{\color{cyan}\textbf{FP:} #1}}
\newcommand{\dcc}[1]{{\color{red}\textbf{DC:} #1}}
\newcommand{\odc}[1]{{\color{orange}\textbf{OD:} #1}}
\newcommand{\abc}[1]{{\color{magenta}\textbf{AB:} #1}}
\newcommand{\rwc}[1]{{\color{green!70!black}\textbf{RW:} #1}}
\newcommand{\cbc}[1]{{\color{magenta!70!black}\textbf{CB:} #1}}

\else
\newcommand{\fpc}[1]{}
\newcommand{\dcc}[1]{}
\newcommand{\odc}[1]{}
\newcommand{\abc}[1]{}
\newcommand{\rwc}[1]{}
\newcommand{\cbc}[1]{}

\fi

\newif\iffinallayout
\finallayoutfalse

\iffinallayout
\newcommand{\finallayout}[1]{#1}
\else
\newcommand{\finallayout}[1]{}
\fi

\usepackage{todonotes}
\usepackage{listings}%
\usetikzlibrary{decorations.pathreplacing}
\usetikzlibrary{fadings}

\title{AlgoNet: $C^\infty$ Smooth Algorithmic Neural Networks}

\author{%
  Felix Petersen\\
  University of Konstanz\\
  \texttt{felix.petersen@uni.kn} \\
  \And
  Christian Borgelt\\
  University of Salzburg\\
  \texttt{christian@borgelt.net} \\
  \And
  Oliver Deussen\\
  University of Konstanz\\
  \texttt{oliver.deussen@uni.kn} \\
}

\begin{document}

\maketitle

\begin{abstract}
Artificial neural networks revolutionized many areas of computer science in recent years since they provide solutions to a number of previously unsolved problems.
On the other hand, for many problems, classic algorithms exist, which typically exceed the accuracy and stability of neural networks.
To combine these two concepts, we present a new kind of neural networks---algorithmic neural networks (AlgoNets).
These networks integrate smooth versions of classic algorithms into the topology of neural networks.
A forward AlgoNet includes algorithmic layers into existing architectures while a backward AlgoNet can solve inverse problems without or with only weak supervision.
In addition, we present the \texttt{algonet} package, a PyTorch based library that includes, inter alia, a smoothly evaluated programming language, a smooth 3D mesh renderer, and smooth sorting algorithms.
\end{abstract}

\section{Introduction}

    Artificial Neural Networks are employed to solve numerous problems, not only in computer science but also in all other natural sciences.
    Yet, the reasoning for the topologies of neural networks seldom reaches beyond empirically based decisions.
    
    In this work, we present a novel approach to designing neural networks---algorithmic neural networks (short: AlgoNet).
    Such networks integrate algorithms as algorithmic layers into the topology of neural networks.
    However, propagating gradients through such algorithms is problematic, because crisp decisions (conditions, maximum, etc.) introduce discontinuities into the loss function. 
    If one passes from one side of a crisp decision to the other, the loss function may change in a non-smooth fashion---it may ``jump''. 
    That is, the loss function suddenly improves (or worsens, depending on the direction) without these changes being locally noticeable anywhere but exactly at these ``jumps''. 
    Hence, a gradient descent based training, regardless of the concrete optimizer, cannot approach these ``jumps'' in a systematic fashion, since neither the loss function nor the gradient provides any information about these ``jumps'' in any place other than exactly the location at which they occur.
    Therefore, a smoothing is necessary, such that information about the direction of improvement becomes exploitable by gradient descent also in the area surrounding the ``jump''. 
    I.e., by smoothing, e.g., an \texttt{if}, one can smoothly, by gradient descent, undergo a transition between the two crisp cases using only local gradient information.

    Generally, for end-to-end trainable neural network systems, all components should at least be $C^0$ smooth, i.e., continuous, to avoid ``jumps''.
    However, having $C^k$ smooth, i.e., $k$ times differentiable and then still continuous components with $k\geq1$ is favorable.
    This property of higher smoothness allows for higher order derivatives and thus prevents unexpected behavior of the gradients.
    Hence, we designed smooth approximations to basic algorithms where the functions representing the algorithms are ideally $C^\infty$ smooth.
    For that, we designed pre-programmed neural networks (restricted to smooth components) with the structure of given algorithms.

    Algorithmic layers can solve sub-problems of the given problem, act as a custom algorithmic loss, or assist in finding an appropriate solution for (ill-posed) inverse problems.
    Such algorithmic losses can impose constraints on predicted solutions through optimization with respect to the algorithmic loss.
    Ill-posed problems are a natural application for algorithmic losses and algorithmic layers.
    
    In this work, we describe the basic concept of algorithmic layers and present some applications.
    In Sec.~\ref{subsec:while}, we start by showing that any algorithm, which can be emulated by a Turing machine, can be approximated by a $C^\infty$ smooth function.
    In Sec.~\ref{sec:impl}, we present some algorithmic layers that we designed to solve underlying problems.
    All described algorithmic layers are provided in the \texttt{algonet} package, a PyTorch \cite{Paszke2017} based library for AlgoNets.

\section{Related Work}

    Related work 
    \cite{tensorflow2015-whitepaper, Che2018, Henderson2018}
    in neural networks focused on dealing with crisp decisions by passing through gradients for the alternatives of the decisions.
    There is no smooth transition between the alternatives, which introduces discontinuities in the loss function that hinder learning, which of the alternatives should be chosen.
    TensorFlow contains a sorting layer (\texttt{tf.sort}) as well as a while loop construct (\texttt{tf.while\_loop}).
    Since the sorting layer only performs a crisp relocation of the gradients and the while loop has a crisp exit condition, there is no gradient with respect to the conditions in these layers.
    Concurrently, we present a smooth sorting layer in Sec.~\ref{subsec:softsort} and a smooth while loop in Sec.~\ref{subsec:while}.
    Theoretical work by DeMillo \textit{et al.} \cite{DeMillo1993} proved that any program could be modeled by a smooth function.
    Consecutive works \cite{Nesterov2005, Chaudhuri2011, Yang2017} provided approaches for smoothing programs using, inter alia, Gaussian smoothing \cite{Chaudhuri2011, Yang2017}.

\section{AlgoNet}

    To introduce algorithmic layers, we show that smooth approximations for any Turing computable algorithm exist and explain two flavors of AlgoNets: forward and backward AlgoNets.

\subsection{Smooth algorithms}

    To design a smooth algorithm, all discrete cases (e.g., conditions of \texttt{if} statements or loops) have to be replaced by continuous or smooth functions.
    The essential property is that the implementation is differentiable with respect to all internal choices and does not---as in previous work---only carry the gradients through the algorithm.
    For example, an \texttt{if} statement can be replaced by a sigmoid-weighted sum of both cases.
    By using a smooth sigmoid function, the statement is smoothly interpreted.
    Hence, the gradient descent method can influence the condition to hold if the content of the \texttt{then} case reduces the loss and influence the condition to fail if the loss is lower when the \texttt{else} case is executed.
    Thus, the partial derivative with respect to a neuron is computed because the neuron is used in the \texttt{if} statement.
    In contrast, when propagating back the gradient of the \texttt{then} or the \texttt{else} case depending on the value of the condition, there is a discontinuity at the points where the value of the condition changes and the partial derivative of the neuron in the condition equals zero.
    
    \noindent
    \begin{minipage}[c]{.5\textwidth}
        \begin{equation}
            s_1(x, s) = \frac{1}{1+e^{-x\cdot s}} \qquad \mbox{with }s=1 \label{math:sigmoid}
        \end{equation}
    \end{minipage}%
    \begin{minipage}[c]{.5\textwidth}
        \begin{equation}
            s_2(x) = \begin{cases}
            0 & \mbox{if }x < 0\\
            1 & \mbox{else}
            \end{cases}\label{math:heaviside}
        \end{equation} 
    \end{minipage}%

    Here, the logistic sigmoid function (Eq.~\ref{math:sigmoid}) is a $C^\infty$ smooth replacement for the Heaviside sigmoid function (Eq.~\ref{math:heaviside}), which is equivalent to the \texttt{if} statement.
    Alternatively, one could use other sigmoid functions, e.g., the $C^1$ smooth step function $x^2 - 2\cdot x^3$ for $x \in [0, 1]$, and $0$ and $1$ for all values before and after the given range, respectively.

    Another example is the max-operator, which, in neural networks, is commonly replaced by the SoftMax operator 
    $\left(\mathrm{SoftMax}(\mathbf{x})_i = \frac{\exp({x_i})}{\sum_{j} \exp({x_j})}\right)$.\\
    After designing an algorithmic layer, we can use it to build a neural network as shall be described in Sec.~\ref{subsec:forwardA} and \ref{subsec:backwardA}.

\subsubsection{Smooth WHILE-Programs}\label{subsec:while}

    In this section, we present smooth and differentiable approximations to an elementary programming language based on the WHILE-language by Uwe Schöning \cite{schoening2008}. 
    The WHILE-language is Turing-complete, and for variables (\texttt{var}) its grammar can be defined as follows:
    \vspace{-.25em}
    \begin{verbatim}
    prog = WHILE var != 0 DO prog END
         | prog prog
         | var := var               //  left and right var unequal
         | var := var + 1           //  left and right var equal
         | var := var - 1           //  left and right var equal     \end{verbatim}
    \vspace{-.25em}
    Here, $\texttt{var}\in \{\texttt{xn}\ |\ \texttt{n}\in\mathbb{N}_0\}$ and while \texttt{x0} is the output, \texttt{xn} where $\texttt{n}\in\mathbb{N}_0$ are variables initialized to $0$ if not set to an input value.
    Thus, $\{\texttt{xn}\ |\ \texttt{n}\in\mathbb{N}_+\}$ are inputs and/or local variables used in the computations.
    Although this interpretation allows only for a single scalar output value (\texttt{x0}), an arbitrary number of output values can be reached, either by interpreting additional variables as output or by using multiple WHILE-programs (one for each output value).
    Provided Church's thesis holds, this language covers all effectively calculable functions.
    The WHILE-language is equivalent to register machines which have no WHILE, but instead IF and GOTO statements.
    
    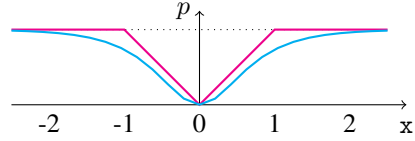
\begin{wrapfigure}{r}{0.4\textwidth}
        \centering
        
        \vspace{-1em}
        
        \begin{tikzpicture}
        \draw[->] (-2.5,0) -- (2.75,0) node[anchor=north, yshift=-1mm] {\texttt{x}};
        \draw	(0,0) node[anchor=north] {0}
        (1,0) node[anchor=north] {1}
        (2,0) node[anchor=north] {2}
        (-1,0) node[anchor=north] {-1}
        (-2,0) node[anchor=north] {-2};

        \draw[->] (0,0) -- (0,1.25) node[anchor=east] {$p$};
        \draw[dotted] (-2,1) -- (2,1);

        \draw[thick, magenta] (-2.5,1) -- (-1,1) -- (0,0) -- (1,1) -- (2.5,1);
        \draw[thick, cyan]   plot [domain=-2.5:2.5] (\x, {1 - (1 / (cosh(2*\x) + 0.001))});

        \end{tikzpicture}
        
        \caption{
            $\phi_0$ (magenta) and $\phi_\infty$ (cyan).%
            \vspace{-1em}
        }

        \label{fig:phiplots}
        
    \end{wrapfigure}
    
    We generate the approximation to this language by executing all statements only to the extent of their probability.
    I.e., we keep track of a probability $p$, indicating whether the current statement is still executed.
    Initially, $p=1$.
    In the body of a while loop, the probability for an execution is \hbox{$p_{k}^{\rm (new)}(x) = p_{k}^{\rm (old)} \cdot \phi_{k}(x) $} where $k$ defines the smoothness of the probability function for exiting the loop $\phi_{k}$.
    To obtain $C^\infty$ smooth WHILE-programs, we used $\phi_{\infty}:\mathbb{R}\rightarrow[0,1]:x\mapsto \frac{\left(e^{s x} - 1\right)^2}{e^{2 s x} + 1} = 1-\mathrm{sech}(s x)$.
    For $C^0$ smoothness, we used the shouldered fuzzy set $\phi_{0}:\mathbb{R}\rightarrow[0,1]:x\mapsto 1 -\max(0,1-|x|) = \min(1,|x|)$.
    For \texttt{x0}...\texttt{xn} initialized as integers, using $\phi_{0}$, the result for the $C^0$ WHILE-program always equals the result for the discrete WHILE-program since the probability is always either 1 or 0.
    For all $k\in\mathbb{N}_0$, $\phi_{k}$ exists.
    Fig.~\ref{fig:phiplots} shows how these exit probability functions behave.
    Because of the symmetry of $\phi$, w.l.o.g., we can assume that $\forall \texttt{x} \in \mathbb{R}_{\geq0}$.
    If the loop (in the discrete version) increases \texttt{x} by 1, \texttt{x} will diverge.
    If \texttt{x} decreases by 1, $p$ converges to 0.
    Since $\phi_0(\texttt{x}) \geq \phi_\infty(\texttt{x})$, it suffices to show that $p$ converges to zero for $\phi_0$.
    Since $p^{\rm (new)}(\texttt{x}) = p^{\rm (old)}(\texttt{x}) \cdot \phi_0(\texttt{x}) \leq \phi_0(\texttt{x}) \leq |\texttt{x}|$, $\texttt{x} := \texttt{x} - p^{\rm (new)}(\texttt{x}) \geq \texttt{x} - |\texttt{x}| = 0$.
    Thus, $\texttt{x}$ and $\phi_0(\texttt{x})$ monotonically decrease and $\texttt{x}\geq 0$.
    Eventually, $\phi_0(\texttt{x}) < 1$. 
    Thus, $p(\texttt{x})\leq \left(\phi_0(\texttt{x})\right)^n \xrightarrow{n\to\infty} 0$.

    Here, $x$ is the value of the current variable, $s$ is the steepness, and $p$ is the probability of the execution.
    To apply the probabilities on the assignment, increment and decrement operators, we redefine them as:
    \vspace{-1.em}
    \begin{align}
        \texttt{x0} := \texttt{x1} \qquad\xrightarrow{} \quad& \texttt{x0} := p \cdot \texttt{x1} + (1-p) \cdot \texttt{x0}\\
        \texttt{x0} := \texttt{x0} + 1 \qquad\xrightarrow{} \quad& \texttt{x0} := \texttt{x0} + p\\
        \texttt{x0} := \texttt{x0} - 1 \qquad\xrightarrow{} \quad& \texttt{x0} := \texttt{x0} - p
    \end{align}
    \vspace{-1.em} \\
    Since the hyperbolic secant is $C^\infty$ smooth, our version of the WHILE-language is $C^\infty$ smooth.
    While the probability converges to zero, it (in most cases) never reaches zero, and the loop would never exit.
    Thus, we introduce $\epsilon>0$ and exit the loop if $p\leq\epsilon$ or a maximum number of iterations is reached.
    Although this introduces discontinuities, by choosing an $\epsilon$ of numerically negligible size, the discontinuities also become numerically negligible.
    \\
    As an experiment, we implemented the multiplication on positive integers, as shown on the left:
    \vspace{-.25em}
\begin{lstlisting}
WHILE x2 != 0 DO       | WITH $p_1 := 1$; $p_1' := p_1 \cdot \phi(\texttt{x2})$ DO
    x3 := x1           |     x3 := $p_1 \cdot \texttt{x1} + (1-p_1) \cdot \texttt{x3}$
    WHILE x3 != 0 DO   |     WITH $p_2 := p_1$; $p_2' := p_2 \cdot \phi(\texttt{x3})$ DO
        x0 := x0 + 1   |         x0 := x0 + $p_2$
        x3 := x3 - 1   |         x3 := x3 - $p_2$
    END                |     WHILE $p_2 \geq \epsilon$
    x2 := x2 - 1       |     x2 := x2 - $p_1$
END                    | WHILE $p_1 \geq \epsilon$\end{lstlisting}
\vspace{-.5em}
    Contrary to the discrete implementation, the smooth interpretation (as on the right) can interpolate the result for arbitrary values $\texttt{x1}, \texttt{x2} \in \mathbb{R}_+$.
    
    Since the WHILE-language is Turing-complete, any high-level program could, in principle, using an appropriate compiler, be translated into an equivalent program in WHILE-language.
    To this WHILE-program, automatic smoothing could be applied using the rules that are illustrated here.
    Of course, there are better ways of smoothing: manual smoothing using domain-specific knowledge and smoothing using a higher-level language outperform the low-level automatic smoothing.
    For example, in a higher-level language, the multiplication would be implemented since it is smooth itself.
    Using domain-specific knowledge, algorithms could be reformulated in such a way that smoothing is possible in a more canonical way.
    To translate a WHILE-program into a neural network, the WHILE loops are considered as recurrent sub-networks.

    \begin{figure}[t]
        \centering
        \vspace{-.025\textwidth}
        \begin{tabular}{ccc}
            \hspace*{-.8cm}
            \subcaptionbox{$\mathtt{x1}\cdot\mathtt{x2}$ }{\includegraphics[width=.35\textwidth]{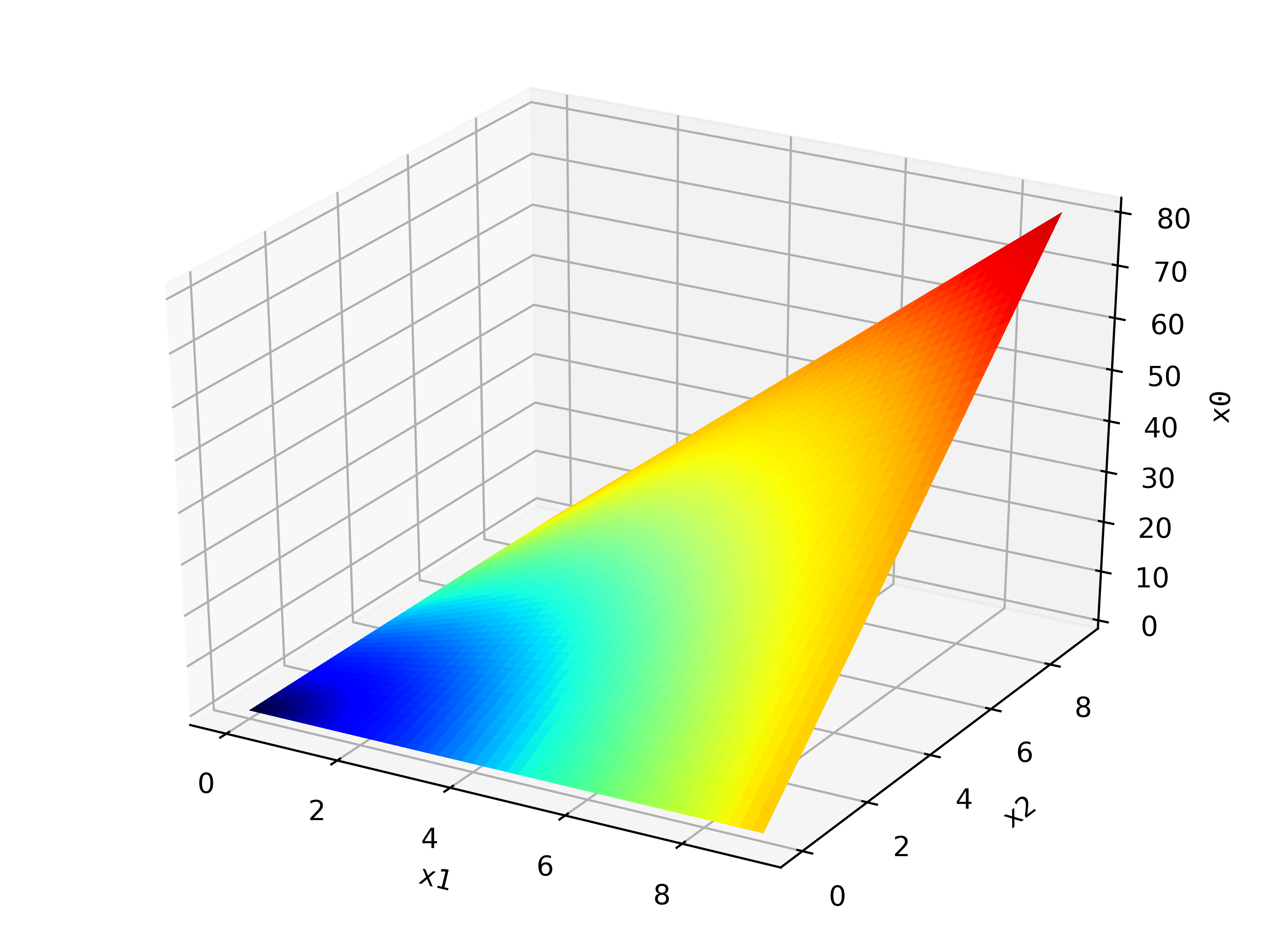}}\hspace*{-.8cm} &
            \subcaptionbox{$C^\infty$ WHILE($\mathtt{x1}$, $\mathtt{x2}$) }{\includegraphics[width=.35\textwidth]{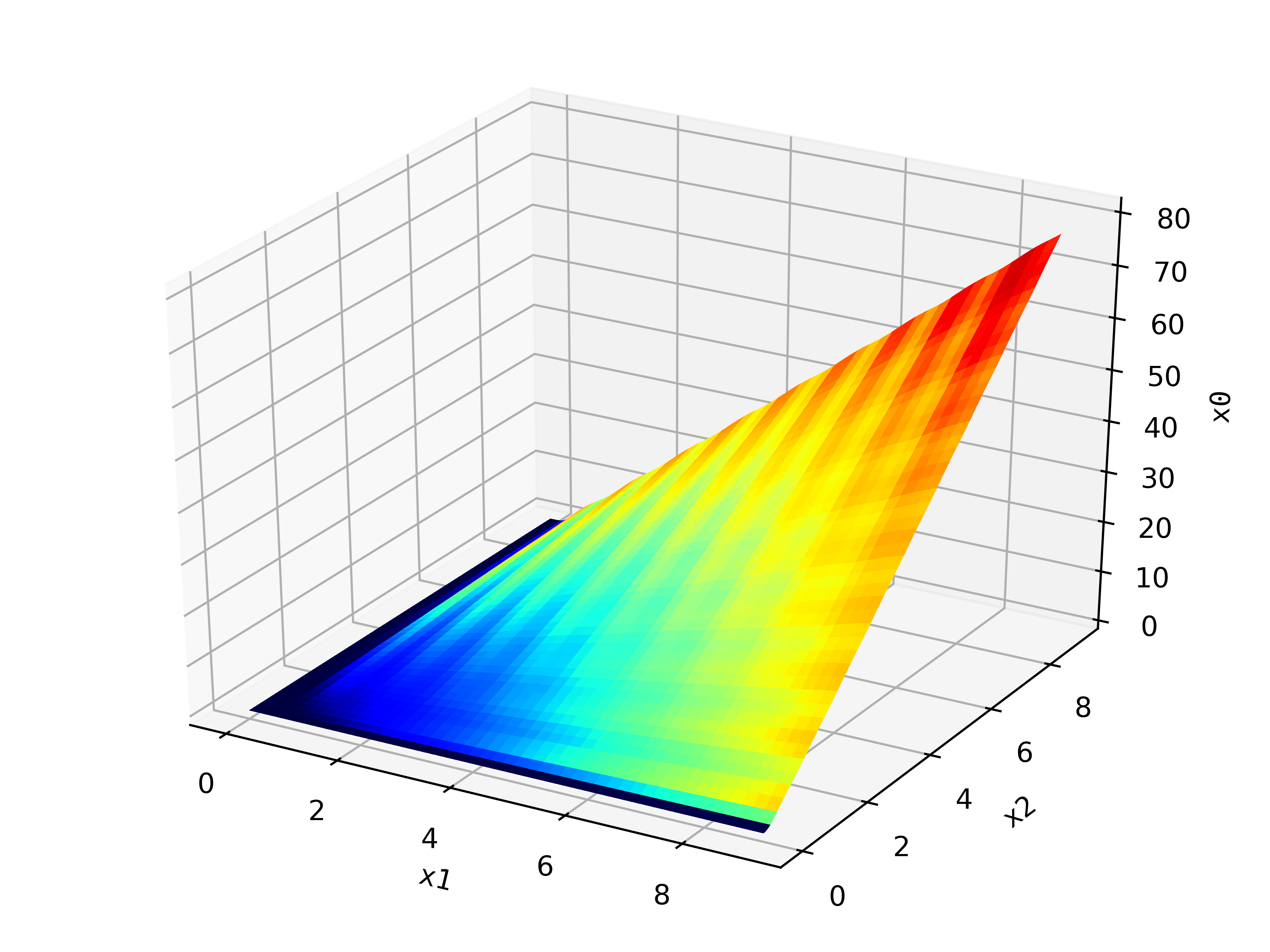}}\hspace*{-.8cm} &
            \subcaptionbox{$C^0$ WHILE($\mathtt{x1}$, $\mathtt{x2}$) }{\includegraphics[width=.35\textwidth]{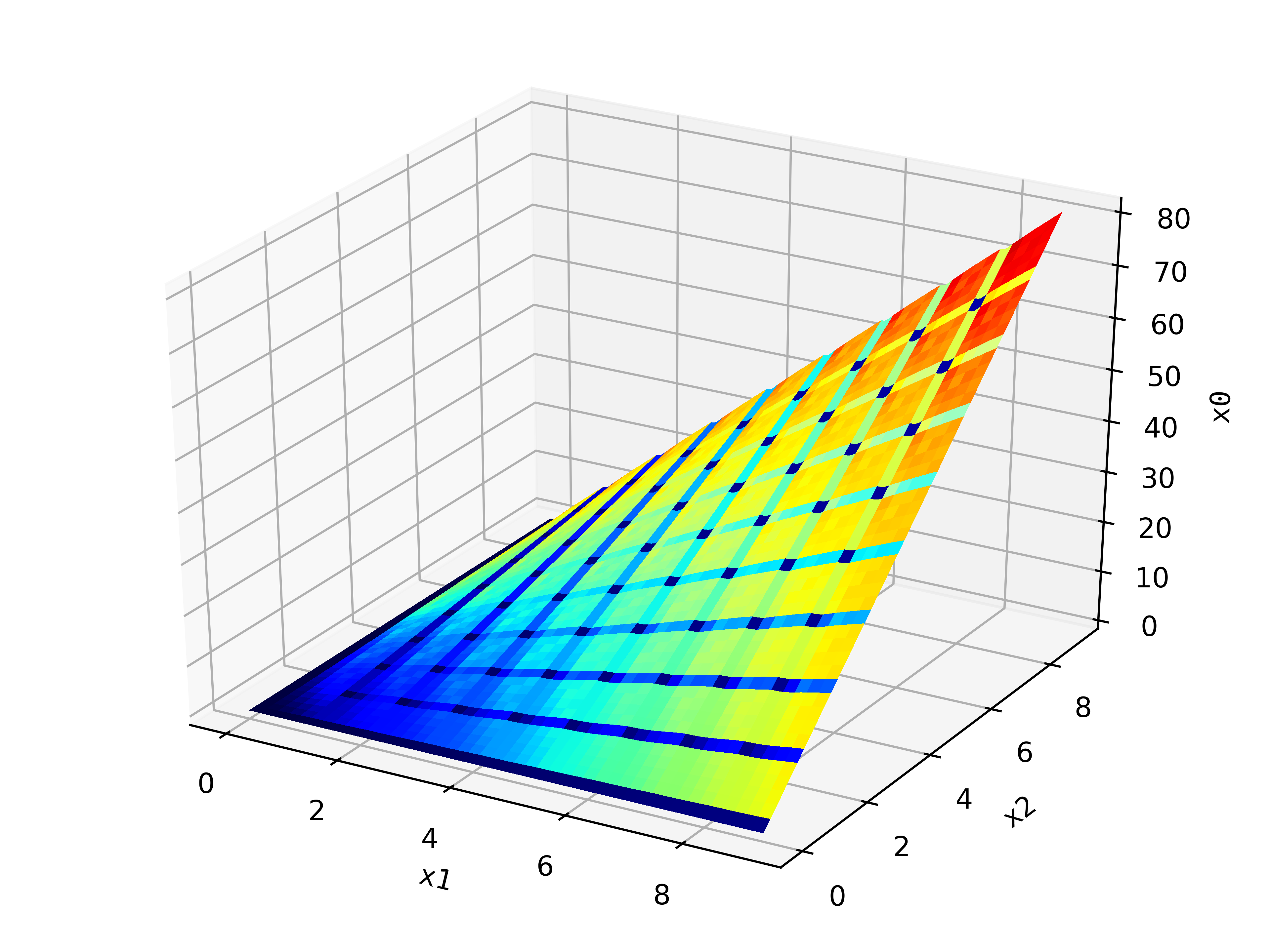}}\hspace*{-.8cm}
        \end{tabular}
        \caption{
            Multiplication of \texttt{x1} and \texttt{x2} implemented as a general multiplication, $C^\infty$ smooth WHILE-program and $C^0$ smooth WHILE-program.
            The height indicates the result of the function.
            Contrary to the common notion, the color indicates not the values but the analytic gradient of the function.
        }
        \label{fig:while}
    \end{figure}
	
\subsection{Forward AlgoNet}\label{subsec:forwardA}

\begin{figure}[h]
        
        \centering
        
        \vspace{-.5em}

            \begin{tabular}{ccc}
                
                \subcaptionbox{fully applied AlgoNet layer \label{subfig:fullforw}}{
                    \begin{tikzpicture}[shorten >=1pt]
                    
                    \draw[thick] (0,0) -- (4,0);
                    \draw[thick] (0,2) -- (4,2);
                    
                    \draw (0,.75) rectangle (4,1.25) node[pos=.5] {AlgoNet};
                    
                    \foreach \x in {0,1,2,3,4,5,6,7,8,9,10,11}
                    \draw[->] (0.166+\x/3,.75) -- (0.166+\x/3,0);
                    \foreach \x in {0,1,2,3,4,5,6,7,8,9,10,11}
                    \draw[->] (0.166+\x/3,2) -- (0.166+\x/3,1.25);
                    
                    \end{tikzpicture}
                } 
                &
                \subcaptionbox{shortcut AlgoNet layer \label{subfig:shortpartforw}}{
                \begin{tikzpicture}[shorten >=1pt]
                
                \draw[thick] (0,0) -- (4,0);
                \draw[thick] (0,2) -- (4,2);
                
                \draw (2,.75) rectangle (4,1.25) node[pos=.5] {AlgoNet};
                
                \foreach \x in {0,1,2,3,4,5}
                \draw[->] (0.166+\x/3,2) -- (0.166+\x/3,0);
                \foreach \x in {6,7,8,9,10,11}
                \draw[->] (0.166+\x/3,.75) -- (0.166+\x/3,0);
                \foreach \x in {6,7,8,9,10,11}
                \draw[->] (0.166+\x/3,2) -- (0.166+\x/3,1.25);
                
                \end{tikzpicture}
                } 
                &
                \subcaptionbox{residual AlgoNet layer \label{subfig:residforw}}{
                \begin{tikzpicture}[shorten >=1pt]
                
                \draw[thick] (0,0) -- (4,0);
                \draw[thick] (0,2) -- (4,2);
                \draw[] (0,1.6) -- (2,1.6);
                \draw[] (0,.4) -- (2,.4);
                \draw[] (0,1.3) -- (2,1.3);
                \draw[] (0,.7) -- (2,.7);
                
                \draw (2,.75) rectangle (4,1.25) node[pos=.5] {AlgoNet};
                
                \node at (1,1.1) {$\vdots$};
                
                \foreach \x in {0,1,2,3,4,5}
                \draw[->] (0.166+\x/3,.4) -- (0.166+\x/3,0);
                \foreach \x in {0,1,2,3,4,5}
                \draw[->, path fading=none] (0.166+\x/3,.7) -- (0.166+\x/3,.4);
                \foreach \x in {0,1,2,3,4,5}
                \draw[->, path fading=none] (0.166+\x/3,1.6) -- (0.166+\x/3,1.3);
                \foreach \x in {0,1,2,3,4,5}
                \draw[->] (0.166+\x/3,2) -- (0.166+\x/3,1.6);
                \foreach \x in {6,7,8,9,10,11}
                \draw[->] (0.166+\x/3,.75) -- (0.166+\x/3,0);
                \foreach \x in {6,7,8,9,10,11}
                \draw[->] (0.166+\x/3,2) -- (0.166+\x/3,1.25);
                
                \end{tikzpicture}
                }
            \end{tabular}
        \caption{
            Different styles of the forward AlgoNet.
            \vspace{-.25em}
        }
        \label{fig:forward-algonet}
    \end{figure}
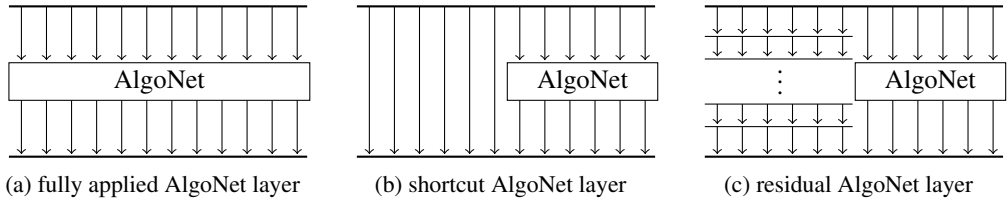

    The AlgoNet can be classified into two flavors, the forward and the backward AlgoNet.
    To create a forward AlgoNet, we use algorithmic layers and insert them into a neural network.
    By doing so, the neural network may or may not find a better local minimum by additionally employing the given algorithm.
    We do so by using one of the following options for each algorithmic layer:
    \begin{itemize}
        \vspace{-.5em}
    	\setlength\itemsep{-.1em}
        \item Insert between two consecutive layers (Fig.~\ref{subfig:fullforw}).
        \item Insert between two consecutive layers and also skip the algorithmic layer (Fig.~\ref{subfig:shortpartforw}).
        \item Add a residual connection and apply the algorithmic layer on the residual part (Fig.~\ref{subfig:residforw}).
        \vspace{-.5em}
    \end{itemize}
    Generally, algorithmic layers do not have trainable weights.
    Regarding the accuracy, the output of $C^\infty$ smooth WHILE-programs differs from the discrete WHILE-programs by a small factor and offset.
    The output of $C^0$ smooth WHILE-programs equals the output of discrete WHILE-programs for integer inputs (discrete WHILE-programs fail for non-integer inputs).
    One could counter this factor and offset for $C^\infty$ smooth WHILE-programs by adding an additional weight and a bias to each assignment in the WHILE-program.
    For that, one should regularize these weights and biases to be close to one and zero, respectively.
    These parameters could be trained on a data set of integer input/output pairs of the respective discrete WHILE-program to fit the algorithmic layer.
    Moreover, the algorithmic layer could also be trained to fit the surrounding layers better.

\subsection{Backward AlgoNet (RAN)}\label{subsec:backwardA}
    
    While forward AlgoNets can use arbitrary smooth algorithms---of course, an algorithm directly related to the problem might perform better---backward AlgoNets use an algorithm that solves the inverse of the given problem.
    E.g., a smooth renderer for 3D-reconstruction, a smooth iterated function system (IFS) for solving the inverse-problem of IFS, and a smooth text-to-speech synthesizer for speech recognition.
    While backward AlgoNets could be used in supervised settings, they are designed for unsupervised or weakly supervised solving of inverse-problems.
    Their concept is the following:\\
    
    \centerline{Input ($\in A$) $\xrightarrow{}$ \texttt{Reconstructor} $\xrightarrow{}$ Goal $\xrightarrow{}$ \texttt{smooth inverse} $\xrightarrow{}$ Smooth version of input ($\in B$)}
    
    This structure is similar to auto-encoders and the encoder-renderer architecture presented by Che \textit{et al.} \cite{Che2018}.
    Such an architecture, however, cannot directly be trained since there is a domain shift between the input domain $A$ and the smooth output domain $B$.
    Thus, we introduce domain translators ($a2b$ and $b2a$) to translate between these two domains.
    Since training with three consecutive components, of which the middle one is highly restrictive, is extremely hard, we introduce a novel training schema for these components: the Reconstructive Adversarial Network (RAN).
    For that, we also include a \texttt{discriminator} to allow for adversarial training of the components $a2b$ and $b2a$.
    Of our five components four are trainable (the \texttt{reconstructor}, the domain translators $a2b$ and $b2a$, and the \texttt{discriminator}), and one is non-trainable (the \texttt{smooth inverse}).
    
\begin{figure*}%
	\centering
	\includegraphics[width=.7\linewidth]{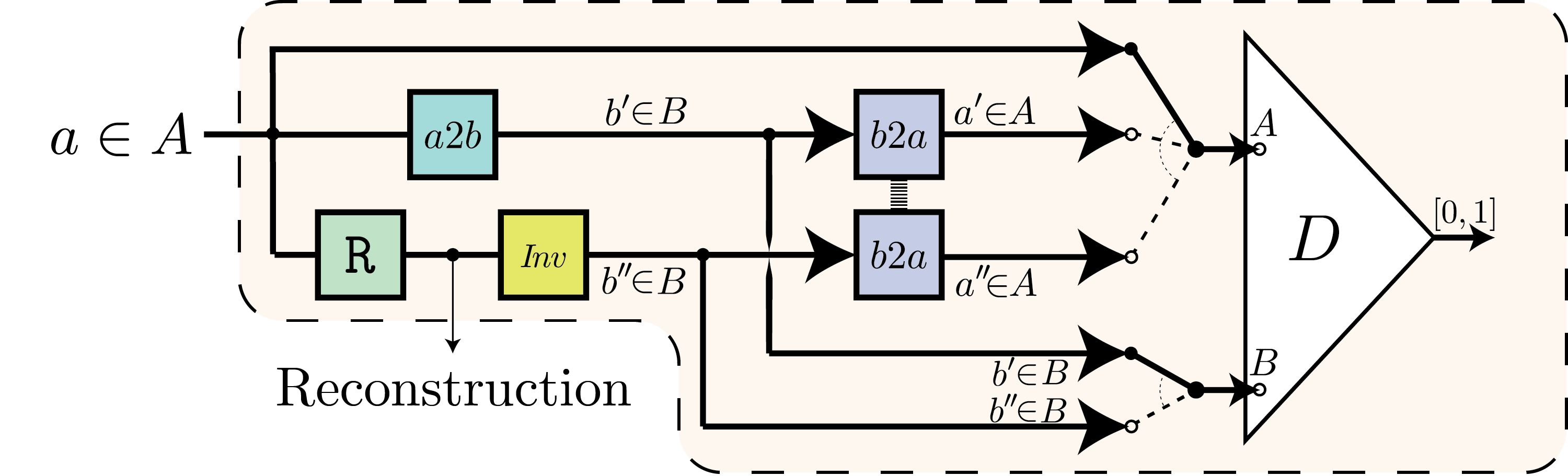}
	\caption{
	RAN System overview. The reconstructor receives an object from the input domain $A$ and predicts the corresponding reconstruction. 
	The reconstruction, then, is validated through our smooth inverse.
	The latter produces objects in a different domain, $B$, which are translated back to the input domain $A$ for training purposes ($b2a$). 
	Unlike in traditional GAN systems, the purpose of our discriminator $D$ is mainly to indicate whether the two inputs match in content, not in style.
	Our novel training scheme trains the whole network via five different data paths, including two which require another domain translator, $a2b$. 
	}
	\label{fig:rancomb}
\end{figure*}

    Since, initially, neither the reconstructor nor the domain translators are trained, we are confronted with a causality dilemma.
    A typical approach for solving such causality dilemmas is to solve the two components coevolutionarily by iteratively applying various influences towards a common solution.
    Fig.~\ref{fig:rancomb} depicts the structure of the RAN, which allows for such a coevolutionary training scheme.

    The discriminator receives two inputs, one from space $A$ and one from space $B$.
    One of these inputs (either $A$ or $B$) receives two values, a real and a fake value; the task of the discriminator is to distinguish between these two, given the other input.
    For training, the discriminator is trained to distinguish between the different path combinations for the generation of inputs.
    Consecutively, the generator modules are trained to fool the discriminator.
    This adversarial game allows training the RAN.

    In the following, we will present this process, as well as its involved losses, in detail.
    Our optimization of \texttt{R}, $a2b$, $b2a$, and $D$ involves adversarial losses, cycle-consistency losses, and regularization losses.
    Specifically, we solve the following optimization:
    \vspace{-.3em}
    \begin{align*}
        \min_{\texttt{R}}\min_{a2b}\min_{b2a}\max_D \ 
        \mathcal{L} 
        \qquad
        \textrm{or in greater detail}
        \qquad \qquad
        \min_{\texttt{R}}\min_{a2b}\min_{b2a}\max_D \ 
        \sum_{i=1}^5\ \left(
        \alpha_i \cdot
        \mathcal{L}_i \right)
        + \mathcal{L}_{\mathrm{reg}}.
    \end{align*} 
    where $\alpha_i$ is a weight in $[0, 1]$ and $\mathcal{L}$, and $\mathcal{L}_i$ shall be defined below.
    $\mathcal{L}_{\mathrm{reg}}$ denotes the regularization losses imposed on the reconstruction output.
    
    We define $b', b'' \in B$ and $a', a'' \in A$ in dependency of $a\in A$ according to Fig.~\ref{fig:rancomb} as
    \begin{align*}
        b' &= a2b(a) & 
        b'' &= \textit{Inv}\circ\texttt{R}(a) & & &
        a' &= b2a(b') &
        a'' &= b2a(b'')
        .
    \end{align*}
    With that, our losses are (without hyper-parameter weights)
    \begin{align*}
        \mathcal{L}_{1} = 
        &\ \mathbb{E}_{a\sim A}[\log D(a, b'')] + \mathbb{E}_{a\sim A}[\log (1 - D(a, b'))] %
        + \ \mathbb{E}_{a\sim A}[\|b'' - b'\|_1] \\
        \mathcal{L}_{2} = 
        &\ \mathbb{E}_{a\sim A}[\log D(a, b'')] + \mathbb{E}_{a\sim A}[\log (1 - D(a'', b''))] %
        + \ \mathbb{E}_{a\sim A}[\|a'' - a\|_1] \\
        \mathcal{L}_{3} = 
        &\ \mathbb{E}_{a\sim A}[\log D(a, b')] + \mathbb{E}_{a\sim A}[\log (1 - D(a'', b'))] %
        + \ \mathbb{E}_{a\sim A}[\|a' - a\|_1] + \mathbb{E}_{a\sim A}[\|b'' - b'\|_1]\\
        \mathcal{L}_{4} = 
        &\ \mathbb{E}_{a\sim A}[\log D(a, b'')] + \mathbb{E}_{a\sim A}[\log (1 - D(a', b''))] %
        + \ \mathbb{E}_{a\sim A}[\|a' - a\|_1] + \mathbb{E}_{a\sim A}[\|b'' - b'\|_1] \\
        \mathcal{L}_{5} = 
        &\ \mathbb{E}_{a\sim A}[\log D(a, b')] + \mathbb{E}_{a\sim A}[\log (1 - D(a', b'))] %
        + \ \mathbb{E}_{a\sim A}[\|a' - a\|_1]
        .
    \end{align*}

    We alternately train the different sections of our network in the following order:
    \begin{enumerate}
        \vspace{-.25em}
        \setlength\itemsep{-.1em}
        \item The discriminator $D$
        \item The translation from $B$ to $A$ ($b2a$)
        \item The components that perform a translation from $A$ to $B$ (\texttt{R}+$Inv$, $a2b$)
        \vspace{-.25em}
    \end{enumerate}
    For each of these sections, we separately train the five losses $\mathcal{L}_1, \mathcal{L}_2, \mathcal{L}_3, \mathcal{L}_4, \textrm{ and } \mathcal{L}_5$.
    In our experiments, we used one Adam optimizer \cite{KingmaB14} for each trainable component (\texttt{R}, $a2b$,~$b2a$,~and~$D$).

\section{Applications}\label{sec:impl}

    In this section, we present specific AlgoNet-layers.
    Specifically, we present a smooth sorting algorithm, a smooth median, a finite differences layer, a weighted SoftMax, smooth iterated function systems, and a smooth 3D mesh renderer.
	
\subsection{SoftSort}
    \label{subsec:softsort}

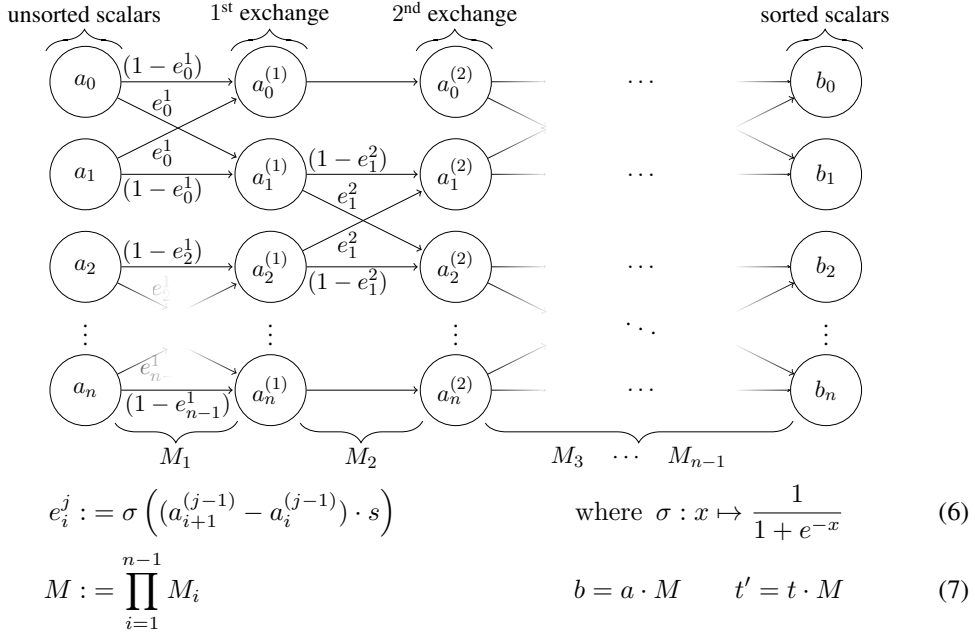
\begin{figure}[h]
        
        \centering
        
        \vspace{-1em}
        \resizebox{.88\textwidth}{!}{
            
            \fontsize{11}{13}
            
            \newcommand{\xddots}{%
                \raise 3pt \hbox {.}
                \mkern 5mu
                \raise 1pt \hbox {.}
                \mkern 5mu
                \raise -1pt \hbox {.}
            }
            
            \newcommand{\exsym}{e}
            
            \centering
            \begin{tikzpicture}[shorten >=1pt]
            \tikzstyle{unit}=[draw,shape=circle,minimum size=1.15cm]
            \tikzstyle{hidden}=[draw,shape=circle,minimum size=1.15cm]
            
            \node[unit](a0) at (0,5){$a_0$};
            \node[unit](a1) at (0,3.5){$a_1$};
            \node[unit](a2) at (0,2){$a_2$};
            \node at (0,1){\vdots};
            \node[unit](an) at (0,0){$a_n$};
            
            \node[unit](a01) at (3,5){$a_0^{(1)}$};
            \node[unit](a11) at (3,3.5){$a_1^{(1)}$};
            \node[unit](a21) at (3,2){$a_2^{(1)}$};
            \node at (3,1){\vdots};
            \node[unit](an1) at (3,0){$a_n^{(1)}$};
            
            \node[unit](a02) at (6,5){$a_0^{(2)}$};
            \node[unit](a12) at (6,3.5){$a_1^{(2)}$};
            \node[unit](a22) at (6,2){$a_2^{(2)}$};
            \node at (6,1){\vdots};
            \node[unit](an2) at (6,0){$a_n^{(2)}$};
            
            \node[unit](b0) at (12,5){$b_0$};
            \node[unit](b1) at (12,3.5){$b_1$};
            \node[unit](b2) at (12,2){$b_2$};
            \node at (12,1){\vdots};
            \node[unit](bn) at (12,0){$b_n$};
            
            \node(d3) at (9,0){$\ldots$};
            \node at (9,1){$\xddots$};
            \node(d2) at (9,2){$\ldots$};
            \node(d1) at (9,3.5){$\ldots$};
            \node(d0) at (9,5){$\ldots$};
            
            \newcommand{\exchangeconnect}[5]{%
                \draw[->] (#1) node [black,xshift=+1.25cm,yshift=+0.25cm]{$(1-#5)$} -- (#3);
                \draw[->] (#1) node [black,xshift=+1.25cm,yshift=-0.35cm]{$#5$} -- (#4);
                \draw[->] (#2) node [black,xshift=+1.25cm,yshift=+0.35cm]{$#5$} -- (#3);
                \draw[->] (#2) node [black,xshift=+1.25cm,yshift=-0.25cm]{$(1-#5)$} -- (#4);
            }

            \exchangeconnect{a0}{a1}{a01}{a11}{\exsym_0^1}{none}
            
            \node(ha2a21) at (1.5,1.25){};
            \node(hanan1) at (1.5,0.75){};
            
            \draw[->] (a2) node [black,xshift=+1.25cm,yshift=+0.25cm]{$(1-\exsym_2^1)$} -- (a21);
            \draw[->,path fading=east] (a2) node [black,xshift=+1.25cm,yshift=-0.35cm]{$\exsym_2^1$} -- (ha2a21);
            \draw[->,path fading=west] (ha2a21) -- (a21);
            
            \draw[->] (an) node [black,xshift=+1.5cm,yshift=-0.25cm]{$(1-\exsym_{n-1}^1)$} -- (an1);
            \draw[->,path fading=east] (an) node [black,xshift=+1.25cm,yshift=+0.35cm]{$\exsym_{n-1}^1$} -- (hanan1);
            \draw[->,path fading=west] (hanan1) -- (an1);

            \draw[->] (a01) -- (a02);
            \exchangeconnect{a11}{a21}{a12}{a22}{\exsym_1^2}{none}
            \draw[->] (an1) -- (an2);

            \draw[->,path fading=east] (a02) -- (7.5, 5.0025);
            \draw[->,path fading=east] (a02) -- (7.5, 4.25);
            \draw[->,path fading=east] (a12) -- (7.5, 3.5);
            \draw[->,path fading=east] (a12) -- (7.5, 4.25);
            \draw[->,path fading=east] (a22) -- (7.5, 2);
            \draw[->,path fading=east] (a22) -- (7.5, 1.25);
            \draw[->,path fading=east] (an2) -- (7.5, .75);
            \draw[->,path fading=east] (an2) -- (7.5, 0);

            \draw[<-,path fading=west] (b0) -- (10.5, 5.0025);
            \draw[<-,path fading=west] (b0) -- (10.5, 4.25);
            \draw[<-,path fading=west] (b1) -- (10.5, 3.5);
            \draw[<-,path fading=west] (b1) -- (10.5, 4.25);
            \draw[<-,path fading=west] (b2) -- (10.5, 2);
            \draw[<-,path fading=west] (b2) -- (10.5, 1.25);
            \draw[<-,path fading=west] (bn) -- (10.5, .75);
            \draw[<-,path fading=west] (bn) -- (10.5, 0);

            \draw [decorate,decoration={brace,amplitude=10pt},xshift=-4pt,yshift=0pt] (-0.5,5.5) -- (0.75,5.5) node [black,midway,yshift=+0.6cm]{unsorted scalars};
            \draw [decorate,decoration={brace,amplitude=10pt},xshift=-4pt,yshift=0pt] (2.5,5.5) -- (3.75,5.5) node [black,midway,yshift=+0.6cm]{$1^{\text{st}}$ exchange};
            \draw [decorate,decoration={brace,amplitude=10pt},xshift=-4pt,yshift=0pt] (5.5,5.5) -- (6.75,5.5) node [black,midway,yshift=+0.6cm]{$2^{\text{nd}}$ exchange};
            \draw [decorate,decoration={brace,amplitude=10pt},xshift=-4pt,yshift=0pt] (11.5,5.5) -- (12.75,5.5) node [black,midway,yshift=+0.6cm]{sorted scalars};

            \draw [decorate,decoration={brace,amplitude=10pt},xshift=-1pt,yshift=0pt] (2.5,-.5) -- (.5,-.5) node [black,midway,yshift=-0.6cm]{$M_1$};
            \draw [decorate,decoration={brace,amplitude=10pt},xshift=-1pt,yshift=0pt] (5.5,-.5) -- (3.5,-.5) node [black,midway,yshift=-0.6cm]{$M_2$};
            \draw [decorate,decoration={brace,amplitude=10pt},xshift=-1pt,yshift=0pt] (11.5,-.5) -- (6.5,-.5) node [black,midway,yshift=-0.6cm]{$M_3\quad \cdots\quad M_{n-1}$};
            
            \end{tikzpicture}
        }
        \vspace{-.5em}
        \begin{align}
            e_i^j :&= \sigma\left( (a_{i+1}^{(j-1)} - a_{i}^{(j-1)}) \cdot s \right) &
            \mathrm{where }\ \  \sigma: x \mapsto \frac{1}{1 + e^{-x}} \\
            M :&= \prod_{i=1}^{n-1} M_i  &
            \qquad b = a \cdot M
            \qquad t' = t \cdot M
        \end{align}
        \vspace{-.5em}
        \caption{
            The structure of SoftSort.
            Here, the exchanges of adjacent elements are represented by matrices $M_i$.
            By multiplying these matrices with tensor $a$, we obtain $b$, the sorted version of $a$.
            By instead multiplying with tensor $t$, we obtain $t'$: $t$ sorted with respect to $a$.
            Using that, we can also sort a tensor with respect to a learned metric.
            For sorting $n$ values, we need $n-1$ steps for an even $n$ and $n$ steps for an odd $n$; to get a probabilistic coarse sorting, even fewer steps may suffice.
            $s$ denotes the steepness of the sorting such that for $s\xrightarrow{}\infty$ we obtain a non-smooth sorting and for infinitely many sorting operations, all resulting values equal the mean of the input tensor.
            In the displayed graph, the two recurrent layers are unrolled in time.
            \vspace{-.5em}
        }

        \label{fig:softsort}
    \end{figure}

    The SoftSort layer is a smooth sorting algorithm that is based on a parallelized version of bubble sort \cite{knuth1998} (see especially Section 5.3.4: Networks for Sorting), which sorts a tensor along an array of scalars by repeatedly exchanging adjacent elements if necessary.
    Fig.~\ref{fig:softsort} shows the structure of the SoftSort algorithm. 
    Contrasting our approach, the sorting layer in TensorFlow \cite{tensorflow2015-whitepaper} is not smooth and does not consider gradients with respect to the ordering induced by the sorting.

\subsection{Finite differences}
    \label{subsec:finitedifferences}
    
    The finite differences method, which was introduced by Lewy \textit{et al.} \cite{Lewy1928}, is an essential tool for finding numerical solutions of partial differential equations. 
    In analogy, the finite differences layer uses finite differences to compute the spatial derivative for one or multiple given dimensions of a tensor.
    For that, we subtract the tensor from itself shifted by one in the given dimension.
    Optionally, we normalize the result by shifting the mean to zero and/or add padding to output an equally sized tensor.
    Thus, it is possible to integrate a spatially- or temporally-derivating layer into neural networks.
    
\subsection{WeightedSoftMax}\label{subsec:weightedsoftmax}

    The weighted SoftMax (short: $\mathfrak{w}$SoftMax) allows a list that is fed to the SoftMax operator to be smoothly sliced by weights indicating which elements are in the list.
    I.e., there are two inputs, the actual values (\textbf{x}) and weights ($w$) from $(0;1]$ indicating which values of \textbf{x} should be considered for the SoftMax.
    Thus, $\mathfrak{w}$SoftMax can be used when the maximum value of values, for which an additional condition also holds, is searched by indicating whether the additional condition holds with weights $w_i\in (0;1]$.
    We define the weighted SoftMax as: 
\begin{align}
    \mathfrak{w}\mathrm{SoftMax}_i(\mathbf{x}, w)
    \ :=\  \frac{\exp ({\mathbf{x}_i}) \cdot w_i}{\sum_{i=0}^{\|w\|-1} \exp ({\mathbf{x}_i}) \cdot w_i}
    &=\  \frac{\exp ({\mathbf{x}_i} + \log w_i) }{\sum_{i=0}^{\|w\|-1} \exp ({\mathbf{x}_i} + \log w_i)} \\
	&=\  \text{SoftMax}_i({\mathbf{x}_i} + \log w_i)\notag
\end{align}
    Accordingly, we define the weighted SoftMin (analogue to SoftMax/SoftMin) as $\mathfrak{w}\mathrm{SoftMin}(\mathbf{x}, w) := \mathfrak{w}\mathrm{SoftMax}(-\mathbf{x}, w)$.
    By that, we enable a smooth selection to apply the SoftMax/SoftMin function only to relevant values.

\subsection{SoftMedian}

The mean is a commonly used measure for reducing tensors, e.g., for normalizing a tensor.
While the median is robust against outliers, the mean is sensitive to all data points.
This has two effects: firstly, the mean is not the most representative value because it is influenced by outliers; secondly, the derivative of a normalization substantially depends on the positions of outliers.
I.e., outliers, which might have accelerated gradients in the first place, can influence all values during a normalization like $x' := x - \bar{x}$.
While one would generally avoid these potentially malicious gradients by cutting the gradients of $\bar{x}$, this is not adequate if changes in $\bar{x}$ are expected.
To reduce the influence of outliers in a smooth way, we propose the SoftMedian, which comes in two styles, a precise and slower as well as a significantly faster version that only discards a fixed number of outliers.

The precise version sorts the tensor with SoftSort and takes the middle value(s).
For that, it is not necessary to carry out the entire SoftSort; instead, only those computations that influence the middle value need to be taken into account.

The faster variant to compute the SoftMedian (of degree $j$) is by its recursive definition in which influence of the minimum and maximum values is reduced as follows:
$\mathrm{SoftMed}(\textbf{x})^{(j)} := 
    \mathfrak{w}\mathrm{SoftMin} \left( 
    \mathfrak{w}\mathrm{SoftMin}(\textbf{x}, \mathrm{SoftMed}(\textbf{x})^{(j-1)}) +
    \mathfrak{w}\mathrm{SoftMax}(\textbf{x}, \mathrm{SoftMed}(\textbf{x})^{(j-1)}), 
    \mathrm{SoftMed}(\textbf{x})^{(j-1)}
    \right)
$
where 
$\mathrm{SoftMed}(\textbf{x})^{(0)} := 1\|\textbf{x}\|$.

\subsection{Smooth Iterated Function Systems}
\label{subsec:smoothIFS}

	Iterated function systems (IFS) allow the construction of various fractals using only a set of parameters.
	For example, pictures of plants like Barnsley`s fern can be generated using only $4\times6=24$ parameters.
	Numerous different plants and objects can be represented using IFS.
	Since IFS are parametric representations, they can be stored in very small space and be adjusted. 
    This can be used, e.g., in a computer game to avoid unnatural uniformity when rendering vegetation by changing the parameters slightly, so that each plant looks slightly different.
	Finally, there are very fast algorithms to generate images from IFS.
	While IFS provide many advantages, solving the inverse problem of IFS, i.e., finding an IFS representation for any given image, is very hard and still unsolved.
    Towards solving this inverse problem, we developed a $C^\infty$ smooth approximation to IFS.
    
	Given a two-dimensional IFS with $n$ bi-linear functions $(f_i)_{i\in\{1..n\}}$ $f_i(x, y) := (x + a_1 + a_2 x + a_3 y, y + a_4 + a_5 x + a_6 y)$, we repeatedly randomly select one function $f\in (f_i)_{i\in\{1..n\}}$ and apply it to an initial position or the proceeding result.
	We do that process arbitrarily often and plot every intermediate step.
    Since it is not meaningful to interpolate multiple functions, because IFS rely on randomized choices, and to provide consistency, we perform these random choices in advance.
	
	The difficulty in this process is the rasterization since no crisp decision correlating pixels to points can be made.
    Thus, we correlate each pixel to each point with a probability $p\in[0;1]$ where $p=1$ is a full correlation and $p=0$ means no correlation at all.
    By applying Gaussian smoothing on the locations of the points, for each point, the probabilities $p\in(0;1)$ for each pixel define the correlation.
    Concluding, for each pixel, the probabilities for all points to lie in the area of that pixel are known.
    By aggregating these probabilities, we achieve a smooth rasterization.
    
    We tested the smooth IFS by optimizing its parameters to fit an image.
    For that, by setting the standard deviation, different levels of details can be optimized.

\subsection{Smooth Renderer}

    Lastly, we include a $C^\infty$ smooth 3D mesh renderer to the AlgoNet library, which projects a triangular mesh onto an image while considering physical properties like perspective and shading.
    Compared to previous differentiable renderers, this renderer is fully and not only locally differentiable.
    Moreover, the continuity of the gradient allows for seamless integration into neural networks by avoiding unexpected behavior altogether.
    By taking the decision which triangles cover a pixel, in analogy to the smooth rasterization in Sec.~\ref{subsec:smoothIFS}, the silhouette of the mesh can be obtained.
    Consecutively, by computing which of these triangles is the closest to the camera smoothly, our renderer's depth buffer is smooth.
    That allows for color handling and shading.
    The smooth depth buffer is the weighted SoftMin of the distance between triangles and camera, weighted with the probability of a triangle to exist at the coordinates of the pixel.
    In experiments, we have used the smooth renderer to perform mesh optimization and mesh prediction using the backward AlgoNet.

\section{Discussion and Conclusion}

    Concluding, in this work, we presented AlgoNets as a new kind of layers for neural networks and developed a $C^\infty$ Turing complete interpreter.
    We have implemented the presented layers on top of PyTorch and will publish our AlgoNet library upon publication of this work.
    Concurrent with their benefits, some AlgoNets can be computationally very expensive.
    For example, the rendering layer requires a huge amount of computation, while the run time of the finite-differences layer is almost negligible.
    On the other hand, the rendering layer is very powerful since it, e.g., allows training a 3D reconstruction without 3D supervision using the backward AlgoNet.
    
    The AlgoNet could also be used in the realm of explainable artificial intelligence \cite{Gilpin2018} by adding residual algorithmic layers into neural networks and then analyzing the neurons of the trained AlgoNet.
    For that, network activation and/or network sensitivity can indicate the relevance of the residual algorithmic layer.
    To compute the network sensitivity of an algorithmic layer, the gradient with respect to additional weights (constant equal to one) in the algorithmic layer could be computed.
    By that, similarities between classic algorithms and the behavior of neural networks could be inferred. 
    An alternative approach would be to gradually replace parts of trained neural networks with algorithmic layers and analyzing the effect on the new model accuracy. 
    
    In future work, we could develop a high-level smooth programming language to improve the smooth representations of higher level programming concepts.
    Adding trainable weights to the algorithmic layers to improve the accuracy of smooth algorithms and/or allow the rest of the network to influence the behavior of the algorithmic layer is subject to future research.
    The similarities of our smooth WHILE-programs to analog as well as quantum computing shall be explored in future work.
    Another future objective is the exploration of neural networks not with a fixed but instead with a smooth topology.

\newpage

{\small%
\printbibliography 
}

\end{document}